\documentclass{article}

\usepackage{microtype}
\usepackage{graphicx}
\usepackage{subfigure}
\usepackage{booktabs} % for professional tables
\usepackage{multirow} % for multirow feature
\usepackage{multicol} % for multicolumn feature

\usepackage{hyperref}
\usepackage{pifont}
\usepackage{float}

\usepackage[accepted]{icml2024}

\usepackage{witharrows}
\usepackage{amsmath}
\usepackage{amssymb}
\usepackage{mathtools}
\usepackage{amsthm}
\usepackage{algorithm}
\usepackage[capitalize,noabbrev]{cleveref}

\theoremstyle{plain}

\theoremstyle{definition}

\theoremstyle{remark}

\usepackage[textsize=tiny]{todonotes}

\icmltitlerunning{Sharpness-Aware Data Generation for Zero-shot Quantization}

\begin{document}

\twocolumn[
\icmltitle{Sharpness-Aware Data Generation for Zero-shot Quantization}



\icmlsetsymbol{equal}{*}

\begin{icmlauthorlist}
\icmlauthor{Hoang Anh Dung}{yyy}
\icmlauthor{Cuong Pham}{yyy}
\icmlauthor{Trung Le}{yyy}
\icmlauthor{Jianfei Cai}{yyy}
\icmlauthor{Thanh-Toan Do}{yyy}
%\icmlauthor{Firstname6 Lastname6}{sch,yyy,comp}
%\icmlauthor{Firstname7 Lastname7}{comp}
%\icmlauthor{Firstname8 Lastname8}{sch}
%\icmlauthor{Firstname8 Lastname8}{yyy,comp}
\end{icmlauthorlist}

\icmlaffiliation{yyy}{Department of Data Science and AI, Monash University, Melbourne, Australia}

\icmlcorrespondingauthor{Hoang Anh Dung}{hoang.dung@monash.edu}
%\icmlcorrespondingauthor{Firstname2 Lastname2}{first2.last2@www.uk}

\icmlkeywords{Machine Learning, ICML}

\vskip 0.3in
]



\printAffiliationsAndNotice{}  % leave blank if no need to mention equal contribution

% Mathbf

\newcommand\norm[1]{\left\lVert#1\right\rVert}

\newcommand{\ba}{\mathbf{a}}
\newcommand{\bb}{\mathbf{b}}
\newcommand{\bc}{\mathbf{c}}
\newcommand{\bd}{\mathbf{d}}
\newcommand{\be}{\mathbf{e}}
\newcommand{\bg}{\mathbf{g}}
\newcommand{\bh}{\mathbf{h}}
\newcommand{\bi}{\mathbf{i}}
\newcommand{\bj}{\mathbf{j}}
\newcommand{\bk}{\mathbf{k}}
\newcommand{\bo}{\mathbf{o}}
\newcommand{\bp}{\mathbf{p}}
\newcommand{\bq}{\mathbf{q}}
\newcommand{\bs}{\mathbf{s}}
\newcommand{\bt}{\mathbf{t}}
\newcommand{\bu}{\mathbf{u}}
\newcommand{\bv}{\mathbf{v}}
\newcommand{\bw}{\mathbf{w}}
\newcommand{\bx}{\mathbf{x}}
\newcommand{\by}{\mathbf{y}}
\newcommand{\bz}{\mathbf{z}}

\newcommand{\bA}{\mathbf{A}}
\newcommand{\bB}{\mathbf{B}}
\newcommand{\bC}{\mathbf{C}}
\newcommand{\bD}{\mathbf{D}}
\newcommand{\bE}{\mathbf{E}}
\newcommand{\bF}{\mathbf{F}}
\newcommand{\bG}{\mathbf{G}}
\newcommand{\bH}{\mathbf{H}}
\newcommand{\bI}{\mathbf{I}}
\newcommand{\bJ}{\mathbf{J}}
\newcommand{\bK}{\mathbf{K}}
\newcommand{\bL}{\mathbf{L}}

\newcommand{\bO}{\mathbf{O}}
\newcommand{\bR}{\mathbf{R}}
\newcommand{\bS}{\mathbf{S}}
\newcommand{\bT}{\mathbf{T}}
\newcommand{\bU}{\mathbf{U}}
\newcommand{\bV}{\mathbf{V}}
\newcommand{\bW}{\mathbf{W}}
\newcommand{\bX}{\mathbf{X}}
\newcommand{\1}{\mathbf{1}}
\newcommand{\bY}{\mathbf{Y}}
\newcommand{\bZ}{\mathbf{Z}}

\newcommand{\ca}{\mathcal{a}}
\newcommand{\cb}{\mathcal{b}}
\newcommand{\cc}{\mathcal{c}}
\newcommand{\ck}{\mathcal{k}}
\newcommand{\co}{\mathcal{o}}
\newcommand{\cp}{\mathcal{p}}
\newcommand{\cq}{\mathcal{q}}
\newcommand{\cs}{\mathcal{s}}
\newcommand{\ct}{\mathcal{t}}
\newcommand{\cu}{\mathcal{u}}
\newcommand{\cv}{\mathcal{v}}
\newcommand{\cw}{\mathcal{w}}
\newcommand{\cx}{\mathcal{x}}
\newcommand{\cy}{\mathcal{y}}
\newcommand{\cz}{\mathcal{z}}

\newcommand{\cA}{\mathcal{A}}
\newcommand{\cB}{\mathcal{B}}
\newcommand{\cC}{\mathcal{C}}
\newcommand{\cD}{\mathcal{D}}
\newcommand{\cE}{\mathcal{E}}
\newcommand{\cF}{\mathcal{F}}
\newcommand{\cG}{\mathcal{G}}
\newcommand{\cH}{\mathcal{H}}
\newcommand{\cI}{\mathcal{I}}
\newcommand{\cJ}{\mathcal{J}}
\newcommand{\cK}{\mathcal{K}}
\newcommand{\cL}{\mathcal{L}}
\newcommand{\cM}{\mathcal{M}}
\newcommand{\cN}{\mathcal{N}}

\newcommand{\cO}{\mathcal{O}}
\newcommand{\cP}{\mathcal{P}}
\newcommand{\cR}{\mathcal{R}}
\newcommand{\cS}{\mathcal{S}}
\newcommand{\cT}{\mathcal{T}}
\newcommand{\cU}{\mathcal{U}}
\newcommand{\cV}{\mathcal{V}}
\newcommand{\cW}{\mathcal{W}}
\newcommand{\cX}{\mathcal{X}}
\newcommand{\cY}{\mathcal{Y}}
\newcommand{\cZ}{\mathcal{Z}}

\newcommand{\ta}{\mathtt{a}}
\newcommand{\tb}{\mathtt{b}}
\newcommand{\tc}{\mathtt{c}}
\newcommand{\td}{\mathtt{d}}
\newcommand{\te}{\mathtt{e}}
\newcommand{\tf}{\mathtt{f}}
\newcommand{\tg}{\mathtt{g}}
\newcommand{\ti}{\mathtt{i}}
\newcommand{\tj}{\mathtt{j}}
\newcommand{\tk}{\mathtt{k}}
\newcommand{\tp}{\mathtt{p}}
\newcommand{\tq}{\mathtt{q}}
\newcommand{\ts}{\mathtt{s}}
\newcommand{\tu}{\mathtt{u}}
\newcommand{\tv}{\mathtt{v}}
\newcommand{\tw}{\mathtt{w}}
\newcommand{\tx}{\mathtt{x}}
\newcommand{\ty}{\mathtt{y}}
\newcommand{\tz}{\mathtt{z}}

\newcommand{\tA}{\mathtt{A}}
\newcommand{\tB}{\mathtt{B}}
\newcommand{\tC}{\mathtt{C}}
\newcommand{\tD}{\mathtt{D}}
\newcommand{\tE}{\mathtt{E}}
\newcommand{\tF}{\mathtt{F}}
\newcommand{\tG}{\mathtt{G}}
\newcommand{\tH}{\mathtt{H}}
\newcommand{\tI}{\mathtt{I}}
\newcommand{\tJ}{\mathtt{J}}
\newcommand{\tK}{\mathtt{K}}
\newcommand{\tL}{\mathtt{L}}

\newcommand{\tO}{\mathtt{O}}
\newcommand{\tP}{\mathtt{P}}
\newcommand{\tR}{\mathtt{R}}
\newcommand{\tS}{\mathtt{S}}
\newcommand{\tT}{\mathtt{T}}
\newcommand{\tU}{\mathtt{U}}
\newcommand{\tV}{\mathtt{V}}
\newcommand{\tW}{\mathtt{W}}
\newcommand{\tX}{\mathtt{X}}
\newcommand{\tY}{\mathtt{Y}}
\newcommand{\tZ}{\mathtt{Z}}

\newcommand{\explain}[2]{\underbrace{#1}_{\parbox{\widthof{#2}}
{\footnotesize\raggedright #2}}}
\newcommand{\dotproduct}[2]{%
  \ooalign{%
    $\m@th#1\circ$\cr
    \hidewidth$\m@th#1\cdot$\hidewidth\cr
  }%
}

\begin{abstract}
\label{sec:abstract}
Zero-shot quantization aims to learn a quantized model from a pre-trained full-precision model with no access to original real training data. The common idea in zero-shot quantization approaches is to generate synthetic data for quantizing the full-precision model. While it is well-known that deep neural networks with low sharpness have better generalization ability, none of the previous zero-shot quantization works considers the sharpness of the quantized model as a criterion for generating training data. This paper introduces a novel methodology that takes into account quantized model sharpness in synthetic data generation to enhance generalization. Specifically, we first demonstrate that sharpness minimization can be attained by maximizing gradient matching between the reconstruction loss gradients computed on synthetic and real validation data, under certain assumptions. We then circumvent
the problem of the gradient matching without real validation
set by approximating it with the gradient matching between
each generated sample and its neighbors. Experimental evaluations on CIFAR-100 and ImageNet datasets demonstrate the superiority of the proposed method over the state-of-the-art techniques in low-bit quantization settings.

\end{abstract}
\section{Introduction}
\label{sec:introduction}
Due to the impressive performance of deep learning models in various fields and applications,  
%In recent years, the world witnessed rapid development of deep learning model and their application in various fields, due to their strong performance over traditional methods. 
there has been great attention to incorporating deep learning models into resource-constrained devices, such as mobiles. 
As a result, optimizing the storage and computational expense of state-of-the-art deep neural networks (DNNs) has become increasingly important. %gathering the attention of the scientific community. 
Among various network compression techniques such as pruning \cite{prunning,channel_pruning}, knowledge distillation \cite{KD,FitNets, DKD,Pham_2024_WACV}, and          quantization~\cite{Courbariaux2015BinaryConnectTD,xnor_net}, network quantization is considered one of the most effective methods. It aims to acquire smaller models with parameters represented in much smaller bit-width (e.g., 1, 2, or 4 bits), yet still achieving competitive performance compared to full-precision (i.e., 32 bits) models %bigger model with 32-bit float parameters
~\cite{Chen2021TowardsMQ,Dong2019HAWQHA,Yang2020FracBitsMP,Pham_2023_WACV,Wei2022QDropRD,Nagel2020UpOD,Courbariaux2015BinaryConnectTD}. 

Network quantization approaches can be generally divided into two groups: quantization-aware training (QAT)~\cite{Krishnamoorthi2018QuantizingDC,Esser2019LearnedSS,Nagel2021AWP} and post-training quantization (PTQ)~\cite{Nahshan2019LossAP,Banner2018ACIQAC,Li2021BRECQPT,Nagel2020UpOD}. While QAT approaches %focus on training the quantized model from scratch and 
have shown outstanding performance comparable to full-precision models, they often require access to a large amount of real training data % that have been used to train the full-precision model} 
and a significant amount of training time. 
On the other hand, post-training quantization (PTQ) approaches only require a small amount of original data and focus on distilling knowledge from a full precision model to a quantized one. %This setting is very realistic due to the shortage of data in many real world scenarios. 

As an attempt to combat a more challenging scenario where there is no access to any part of the original data (e.g., due to privacy), zero-shot quantization (ZSQ) setting that does not require any original data has been proposed. The common idea in ZSQ~\cite{qimera,Genie,HAST}  is to generate a set of synthetic data %from scratches 
%to be beneficial 
as a calibration set for the quantization process, by taking advantage of information from a full-precision model. 
Most of the prominent ZSQ works utilize the batch normalization (BN) statistics information in BN layers of the full-precision model, i.e., they try to generate synthetic samples such that the feature distribution of generated samples matches the BN statistics~\cite{qimera,ZeroQ}. %Unfortunately, this approach can result in homogeneous sample generation with limited information due to the lack of diversity factor. As the result, investigating the diversity of generated data has also gather attention of some recent works, which take into account the inter and intra-relationship of synthetic samples based on their labels~\cite{IntraQ,DSG}. 
Some other methods pay attention to the boundary information of the full-precision model and try to generate data that are near the decision boundary of the full-precision model~\cite{qimera,HAST,AdaDFQ}. 
Although different criteria have been investigated when generating synthetic samples in ZSQ~\cite{qimera,Genie,HAST,AdaDFQ}, 
%\blue{most of them %are just intuitive-based heuristic approaches that 
%often lack quantifiable metrics for estimating the impacts of the synthetic dataset on the generalization of the quantized model calibrated on it.} 
%They do not theoretically justify connections of the generated data and the model's generalization. 
none of them considers connections between the generated data and the model's sharpness, despite that it has been shown reducing the model's sharpness can improve its generalization~\cite{FirstSam,ESAM,LookSam}. 
%\blue{Lacking of consideration of the generalization could cause quantized model to be overfitted on the generated data, which consequently cause the poor performance of the real testing data.}  %Moreover, investigation about other aspects of synthetic data such as their impact on model's generalization is very limited, which consequently lead to overfitting problem in most  ZSQ and PTQ methods due to the lack of data. 

To this end, in this paper, we propose a novel method for ZSQ that takes into account the impacts of the generated data on the sharpness of the quantized model during the generation process. 
%the sharpness of the quantized model when generating data %, as a metric to evaluate the model's generalization. 
Specifically, we aim to generate a set of training images such that using these generated images for learning the quantized model will not only result in a good knowledge transfer from the full-precision model %to the quantized model 
but also minimize the sharpness of the quantized model when evaluating on a validation set of real images. 
However, the real validation set is not available in the context of ZSQ. To overcome this challenge, we rigorously show that under some assumptions, the sharpness minimization can be achieved by maximizing the matching between the gradients of the reconstruction loss %w.r.t. quantized models when 
evaluated on the generated data and real validation data, respectively. We then circumvent the problem of the gradient matching without real validation set by approximating it with the gradient matching between each generated sample and its neighbors, which can be done through an SAM-like optimization.
%propose a novel approach to approximate the gradient matching without the need for the real validation set. Specifically, we propose to generate samples that are diverse and for each sample, the gradients w.r.t. the quantized model when evaluated at the sample and its neighbors are encouraged to be matched, which can be done through a SAM-like optimization. 
The contributions of this work can be summarized: 
%To our best knowledge, this paper is the first work that theoretically leverages the Sharpness-Aware Minimization (SAM) as a criteria for generating the training data in the ZSQ problem.  
 %In summary, our contributions are:
 
%we propose a new zero-shot data generation method, that take into account their impact on quantized model's sharpness information, an effort to improve the quantized model's generalization. Additionally, we try to derive and link the sharpness information of the model to the gradient matching of artificial samples, and propose an efficient method from this. 
%To our best knowledge, there is no existing work that try to connect the relationship between the impact of synthetic data generation and their gradient matching to the model's sharpness in zero-shot quantization setting. In summary, our contribution includes:

  \ding{182 } To our knowledge, this paper is the first one that %theoretically 
  rigorously leverages the Sharpness-Aware Minimization (SAM) as a criterion for generating the training data in the ZSQ problem.\\  %
  %We propose a novel ZSQ method in which the SAM is taken into account as a criterion when generating the data, i.e., the data is generated such that it encourages the flatness of the quantized model. %Consequently it enhances the model's generalization. 
  %The proposal of a novel zero-shot data generation methods that studies the impact of generated data to the model's sharpness after training.
  \ding{183} We %theoretically 
  link the model sharpness with the gradient matching, i.e., maximization of the matching of the gradients of the reconstruction loss w.r.t. model parameters %model when evaluated 
  on the generated and validation data leads to sharpness reduction.\\
  \ding{184} We propose a novel approach to approximate gradient matching without the need for a real validation set.\\
  %\item Our method outperforms the state-of-the-art ZSQ methods under very low-bit quantization settings. 
  %\item We show the connection between gradient matching of artificial data and the sharpness information through theoretical derivation under some assumptions, and design an efficient mechanism exploiting this connection to reduce computational expense.
  \ding{185} Experimental results demonstrate that our novel Sharpness-Aware Data Generation (SADAG) method outperforms the state-of-the-art ZSQ methods under low-bit quantization settings. 

\vspace{0.2cm}
\section{Related work}
\label{sec:related_work}
\subsection{Uniform quantisation}
\label{subsec:quantization_scheme}
%Network quantization is a family of techniques to reduce memory storage and to accelerate model inference by representing full-precision weights with low-bit integers. 
Uniform quantization is the most popular quantization technique 
for quantizing DNNs, %to reduce memory storage and to accelerate model inference, 
due to its simplicity. The integer weight of a uniformly quantized model can be determined by the quantizer $Q_b$ as:
\begin{equation}
    \hat{w} = Q_b(w;s) = s \times \operatorname{clip} \left( \left\lfloor \frac{w}{s} \right\rceil, n, p \right),
\end{equation}
where \(s\) represents the scaling factor, $\lfloor . \rceil$ denotes the rounding-to-nearest function, and clip() represents the clipping function. For instance, to represent a quantized model with unsigned $b$ bits, we can have $n=0$, $p=2^{b-1}$, while $s = \frac{max(w)-min(w)}{2^b-1}$.
The recent state-of-the-art post training quantization (PTQ) approaches~\cite{Wei2022QDropRD,Genie} have adopted adaptive rounding~\cite{Nagel2020UpOD} to improve the performance of uniform quantization further: 
\begin{equation}
    \hat{w} = s \times \operatorname{clip} \left( \left\lfloor \frac{w}{s} \right\rfloor + h(v), n, p \right) 
\end{equation}
where $h(v) \in [0, 1]$ is a learnable function that converges towards either 0 or 1.
In this work, we also adopt %this weight quantization scheme.
the adaptive rounding~\cite{Nagel2020UpOD} for weight quantization. 

\subsection{Zero shot quantization and data generation}
%Instead of training from scratch, 
To address the lack of calibration data, Zero-shot quantization (ZSQ) approaches often aim to exploit information from full-precision models and generate synthetic data that matches that information. The most intuitive method for synthesizing the data is to utilize the cross-entropy (CE) loss between the prediction of the full-precision model and the synthetic label. % $\mathcal{L}_{CE}(x,y)  = E[\mathtt{CE}(x,y)]$. 
However, ZeroQ~\cite{ZeroQ}, one of the early ZSQ works, argues that the distribution mismatch between synthetic data and real data may lead to a significant gap in performance. Their core idea to tackle this problem is to reconstruct synthetic data based on batch normalization (BN) statistics from the BN layers of the full-precision model, which has led to significant performance improvement. 
Since then, built upon this idea, more and more data generation methods for ZSQ have been proposed to incorporate other factors. For instance, KMDFQ~\cite{10132082} and GDFQ~\cite{Xu2020GenerativeLD} assert that ZeroQ~\cite{ZeroQ} disregards the class and distribution attributes inherent in the real dataset. To align the distribution characteristics of between the synthetic and real data, they propose to minimize both the BN loss and the cross-entropy loss of synthetic noise, sampled from predefined distributions. Genie~\cite{Genie} directly constructs samples by optimizing according to BN statistics stored in full-precision models,  %\blue{However, instead of learning from a random noise which the size equals image size, they adopt a simple generator that outputs generated data from an embedding vector initialized from Gaussian distribution, and learn both the generator and the embedding simultaneously.} 
where both the generator and its inputs initialized from a Gaussian distribution are learned simultaneously. 
%They also try to tackle the checkerboard artifacts problem of th traditional data generator model by incorporating an augmentation layer. 
Qimera~\cite{qimera} and HAST~\cite{HAST} both propose to generate boundary-supporting samples that they argue would be important for the quantization process, but with fairly different approaches. While HAST tries to reward samples based on their uncertainty from CE loss, Qimera tries to acquire samples lying within the separating boundary between classes by mixing up class embeddings. 
Other prominent works include DSG~\cite{DSG} and AdaDFQ~\cite{AdaDFQ}. DSG~\cite{DSG} points out the lack of diversity in generated data due to BN statistics optimization. To improve the diversity, they add a margin threshold when minimizing BN statistics mismatch.
%can be considered is  and DSG~\cite{DSG}, which points out the lack of diversity problems in generated data due to BNS optimization, by adding a margin threshold when minimizing BNS mismatch, or 
AdaDFQ ~\cite{AdaDFQ} proposes to combine both boundary and sample diversity optimization. 
{Unfortunately, while different criteria have been proposed in previous works for the data generation process, most of them did not investigate how those criteria link to the model's generalization in a rigorous way. In this paper, we propose a novel method for ZSQ in which the sharpness of the quantized model is directly taken into account when synthesizing the data.}
%\blue{It is worth noting that while there are different criteria are proposed and used when generating data for the ZSQ. None of previous works directly consider the generalization of the quantized model when generating the data, i.e., generating the data such that using those generated data for learning for quantized model would enhance the model generalization. To fill this gap, }  
%Unfortunately, there is still a large gap in performance between the data generated based on BNS and original training data, despite incurring extra computational cost for generation process.
\vspace{0.1cm}
\subsection{Sharpness aware minimization}
Sharpness aware minimization (SAM)~\cite{FirstSam} is an approach proposed to improve the model's generalization. %Proposed by~\cite{FirstSam}, the main purpose of this work aims to combat the problem of rugged surface of the model's loss landscape. 
%\blue{By minimizing the fluctuation in the loss of the model which is add with a weight perturbed noise, SAM can guide the model to a flat local optimum and has shown promising results}. 
By minimizing the loss value and the loss sharpness during training, SAM can guide the model to a flat local optimum, showing promising results. 
Departing from this idea, various variants of SAM have been proposed to address the shortcomings of this approach. In~\cite{LookSam}, the authors propose LookSAM as an attempt to improve efficiency while still retaining the benefits of the SAM algorithm. They claimed that their method is the first to successfully
scale up the batch size when training Vision Transformers with SAM. {Another SAM variant is Efficient SAM~\cite{ESAM}, which attempts to reduce the computational cost of the original SAM. In each iteration of the training, Efficient SAM optimizes the model's sharpness on a set of weights that are stochastically selected. In addition, the sharpness loss is only evaluated on a subset of data that is sensitive to the model's sharpness.}
%\blue{Recently, GSAM~\cite{GSAM} and SAGM~\cite{SAGM} have been proposed to alleviate the gradient conflicts between the perturbed loss and the model's sharpness, aiming to enhance model generalization. }
{One of the problems with SAM is the  gradient conflicts between the perturbed loss and the model's sharpness. Recently,
GSAM~\cite{GSAM} and SAGM~\cite{SAGM} have been proposed to address this gradient conflict problem by modifying the gradient from SAM and removing conflict components.}

% \blue{However, the above methods still inherent the key problem of SAM, i.e. the gradient conflicts between the perturbed loss and the model's sharpness. Recently,
% GSAM~\cite{GSAM} and SAGM~\cite{SAGM} have been proposed to address this issue by modifying the gradient from SAM and removing the conflict component.} %,these methods have shown improvement over vanilla SAM method.} 
%Despite of being extensively investigated its effectiveness in training deep neural network, the impact of data generation in SAM for deep learning has not been investigated. 

\subsection{Sharpness aware minimization for model quantization}
Although there is limited investigation in model generalization, some previous works have attempted to incorporate SAM into %setting of post training 
model quantization. One of the early works is SAQ~\cite{SAQ}, which demonstrates the existence of a sharper
loss landscape in the low-precision model than in the full-precision model. Their work introduces several ways to incorporate sharpness aware loss on optimizing model weights, and proposes an efficient scheme to incorporate SAM without incurring significant cost. 
The recent work Bit-shrinking~\cite{bitshrink} proposes to gradually reduce the number of bits from high bit-width to the target bit-width such that the sharpness of intermediate quantized models does not go over a threshold. 

{Another recent work, ZSAQ~\cite{sam_ptq}, jointly learns the quantized model and the generator with an adversarial learning strategy 
through a minimax optimization.}
%{The generator is trained such as it maximizes the discrepancy between the output distributions of the full-precision model and the quantized model, while the quantized is trained such as it minimizes that discrepancy and reduces the sharpness of the quantized model. From the hard sample perspective, their generator aims to generator hard samples, i.e., the samples that maximizes the discrepancy between the output distributions of the full-precision model and the quantized model, while the quantized is trained on those hard samples with the aims to minimize that discrepancy and the sharpness.}
{It is worth noting that while both ZSAQ~\cite{sam_ptq} and ours have the same objective, i.e., reducing the sharpness of the quantized model, when generating data, their generator does not consider the impact of generated samples on the sharpness of the model, while ours explicitly considers that. In other words, we aim to find a set of synthetic data that is directly beneficial for the model's sharpness over the hidden real data distribution. %In addition, to ensure the generalization of the quantized model, the model's sharpness should be defined on the real data. So we link our generated data with the real data distribution, even with no real data available. 
%we depart with the sharpness evaluated on a real validation set and then link to the generated images. 
Unlike our approach, ZSAQ~\cite{sam_ptq} only evaluates the sharpness of the model on the generated data which may not well generalize to real data.}

%Their approach tried to add perturbation to the weight of quantized model that maximize the discrepancy between the prediction of full precision model and the quantized model over synthetic data from a generator. 

%However, both of these methods provide the looks of optimizing Sharpness property of the model's weight directly, instead of how the synthetic calibrated data impact the sharpness of quantized model's loss landscape.
\section{Method}
\label{sec:method}
\subsection{Problem definition}
Suppose we have a large validation set $X^{(V)}$, and given a deep learning model $\mathtt{f}(.)$ with its pretrained weight $\theta_{FP}$, our ultimate goal is the generation of a small synthetic dataset $X^{(T)}$,  such that using $X^{(T)}$ for learning the quantized model $\theta_{Q}$ will result in a good transfer of knowledge from the full-precision model $\theta_{FP}$ to the quantized model $\theta_{Q}$ on the validation set $X^{(V)}$. 
%\textcolor{red}{{Need to check: Similar to what you mentioned above, we can say like this: We want to generate $X^{(T)}$ such that (i) using $X^{(T)}$ for learning the quantized model $\theta_{Q}$ will result in a good transfer of knowledge from the full-precision model $\theta_{FP}$ to the quantized model $\theta_{Q}$ on the validation set $X^{(V)}$, i.e., $\min \cL_{R}(...)$. }   \textcolor{red}{\textbf{\underline{In addition}}, (ii), we want that the quantized model $\theta_{Q}$ locates in a flat (or not too sharp) region of the (above $\cL_{R}$) loss landscape (for enhancing the generalization).
%Combining (i) and (ii), we will solve:  $\min  \cL_{SAM}(...)$}.}
In the context of model quantization, the transfer knowledge process of a model $\theta_{FP}$ to a quantized model  $\theta_{Q}$ can be done with the layer-wise reconstruction loss on a calibrated set $X$ as:
\begin{equation}
\small
\label{eq:reconstruction_loss_definition}
          \begin{aligned}[b]
                & \cL_{R}( \theta_{Q},\theta_{FP},X) = \frac{1}{2}\sum_{i=1}^{|X|} \sum_{l=1}^{L} \norm{\mathtt{f}(\theta_{FP},\bx_i,l)-\mathtt{f}(\theta_{Q},\bx_i,l)}^{2}, 
          \end{aligned}
    \end{equation}
where $L$ is the number of layers of the model $\mathtt{f}$; %$\theta_{FP}$ and $\theta_{Q}$, while 
$\mathtt{f}(\theta_{FP},\bx_i,l)$ and $\mathtt{f}(\theta_{Q},\bx_i,l)$ are respectively the $l^{th}$-layer outputs of the full-precision and quantized models w.r.t the input sample $\bx_i$; $|.|$ denotes the cardinality of a set. 

In addition, (ii) in order to enhance the generalization of the quantized model $\theta_{Q}$,  we want that $\theta_{Q}$ locates in a flat (or not too sharp) region of the above $\cL_{R}$ loss landscape. 
%To achieve the above two goals, 
%Following~\cite{FirstSam}, we learn the quantized model $\theta_Q$ that minimizes the following 
Following~\cite{FirstSam}, the flatness of the quantized network $\theta_{Q}$ is defined with the 
sharpness-aware (SAM) loss:
\begin{equation}
    \label{eq:L_SAM_definition}
          \begin{aligned}[b]
          & \cL_{SAM}(\theta_{Q},\theta_{FP},X^{(V)}) = \cL_{R}(\theta_{Q} + \epsilon,\theta_{FP},X^{(V)})  \\
          & - \cL_{R}(\theta_{Q},\theta_{FP},X^{(V)}) \\
          & \text{s.t: } 
                \epsilon =  \arg \max_{||\epsilon|| \leq \rho} \cL_{R}(\theta_{Q} + \epsilon,\theta_{FP},X^{(V)}),
          \end{aligned}
    \end{equation}
where $\epsilon$ is a small perpetuation in the neighborhood with radius $\rho$ to the model's weight that increases the loss of the quantized model $\theta_{Q}$ the most.
The closed-form solution for $\epsilon$ is: 
\begin{equation}
    \label{eq:eps_definition}
          \begin{aligned}[b]
          \epsilon=\rho \frac{\triangledown_{\theta_Q} \cL_{R}(X^{(V)})}{||\triangledown_{\theta_{Q}}\cL_{R}(X^{(V)})||},
          \end{aligned}
    \end{equation}
where $\triangledown_{\theta_Q} \cL_{R}(X^{(V)})$ is the short for $ \nabla_{\theta_{Q}}\cL_{R}(\theta_{Q},\theta_{FP},X^{(V)})$, which is the derivative of the reconstruction loss $\cL_{R}$ between the full-precision model $\theta_{FP}$ and the quantized model $\theta_{Q}$.

 In summary, we aim to generate a calibration set $X^{(T)}$ such that a calibration process using $X^{(T)}$ to learn the model $\theta_{Q}$
 can minimize the sum of the two losses in Eqs.  ($\ref{eq:reconstruction_loss_definition}$)\&($\ref{eq:L_SAM_definition})$.
 
 %with eq. ($\ref{eq:reconstruction_loss_definition}$) is equivalent to optimizing model $\theta_{Q}$ on the validation set $X^{(V)}$ with the sum of two losses in eq.  ($\ref{eq:reconstruction_loss_definition}$) and 
 %($\ref{eq:L_SAM_definition})$. This is because in zero-shot setting we do not have access to the validation set $X^{(V)}$.

% Here, \textcolor{red}{similar training behavior} means the two training have similar update trajectories, i.e., their gradients w.r.t models during each update step should be similar.

\subsection{Sharpness-aware data generation}

Suppose that we already initialize a training set $X^{(T)}$. Given the full-precision model $\theta_{FP}$, we start by initializing the quantized model $\theta_{Q}$ without any data, and then calibrate it in the training set by minimizing the reconstruction loss in Eq.  (\ref{eq:reconstruction_loss_definition}) (i.e., using $X^{(T)} $ to update $\theta_{Q}$ with one-step gradient descent to get $\theta^*_{Q}$):
\begin{equation}
\label{eq:quantization_loss}
          \begin{aligned}[b]
                & \theta_{Q}^{*} = \arg \min_{\theta_Q} \cL_{R}( \theta_{Q},\theta_{FP},X^{(T)}). 
          \end{aligned}
    \end{equation}
 Using Eqs. ($\ref{eq:reconstruction_loss_definition}$)\&(\ref{eq:L_SAM_definition}), we can represent the sum of the SAM loss and the reconstruction loss of the %expected 
 calibrated model {$\theta_{Q}^{*}$}:
\begin{equation}
 \small
    \label{eq:L_SAM_expected}
          \begin{aligned}[b] 
          &\qquad \cL_{TOTAL}(\theta_{Q},\theta_{FP},X^{(V)},X^{(T)}) \\
          & = \cL_{SAM}(\theta_{Q}^{*},\theta_{FP},X^{(V)}) + \cL_{R}( \theta_{Q}^{*},\theta_{FP},X^{(V)}) \\
          & = \explain{\cL_{R}(\theta_{Q}^{*} + \epsilon,\theta_{FP},X^{(V)}) - \quad \cL_{R}(\theta_{Q}^{*},\theta_{FP},X^{(V)})}{First term} \\
          & \quad + \explain{\cL_{R}(\theta_{Q}^{*},\theta_{FP},X^{(V)}) - \cL_{R}(\theta_{Q},\theta_{FP},X^{(V)})}{Second term} \\
          & \quad  + \explain{\cL_{R}(\theta_{Q},\theta_{FP},X^{(V)})}{Third term} \\
          & \text{s.t: } 
                \epsilon =  \arg\max_{||\epsilon|| \leq \rho} \cL_{R}(\theta_{Q}^{*} + \epsilon,\theta_{FP},X^{(V)}). 
          \end{aligned}
    \end{equation}

 The synthetic dataset $X^{(T)}$ that optimizes the quantized model's sharpness and performance can be acquired by  minimizing $\cL_{TOTAL}$. 
It is worth noting that the third term $ \cL_{R}(\theta_{Q},\theta_{FP},X^{(V)})$ is independent from the training set $X^{(T)}$ that we want to generate, so we can ignore this term.
\normalsize

\subsection{Gradient matching}
 To simplify the representation, for the rest of the paper, we use  $\triangledown_{\theta} \cL_{R}(X)$ to stand for $\triangledown_{\theta} \cL_{R}(X, \theta,\theta_{FP})$, which represents the gradient of the reconstruction loss between some model $\theta$ with full-precision model $\theta_{FP}$ over some data $X$. %We use this annotation instead of its full form for the rest of the paper.
Let $\delta_{\theta_{Q}} = \theta_{Q}^{*}-\theta_{Q}$  be the difference in weights of the quantized model $\theta_{Q}$  before and after calibrated with the training set $X^{(T)}$. In practice, we only use a single-step gradient descent for the model update, so $\delta_{\theta_{Q}} = -\alpha \triangledown_{\theta_Q} \cL_{R}(X^{(T)})$, where $\alpha$ is the learning rate. % of quantized model.

Using the first order Taylor expansion for $\cL_R(\theta_Q^{*}+\epsilon,\theta_{FP},X^{(V)})$ around $\theta_{Q}^{*}$ in the first term and for $\cL(\theta_Q^{*},\theta_{FP},X^{(V)})$ around $\theta_{Q}$ in the \textit{ second term} of $\cL_{TOTAL}$ in~(\ref{eq:L_SAM_expected}), we have:
%$\cL_{R}(\theta_{Q}^{*} + \epsilon,\theta_{FP},X^{(V)})$ (Eq. \ref{eq:L_SAM_expected}) above at the point $\theta_{Q}^{*}$, we have:
%for the Eq. \ref{eq:L_SAM_expected} above we have:
%\begin{equation}
%        \label{eq:estimate_sam}
%          \begin{aligned}[b]
%                & \qquad \cL_{SAM}( \theta_{Q}^{*},\theta_{FP},X^{(V)}) \\
%                &\approx  \epsilon^{T} \frac{\partial \cL_{R}(\theta_{FP}, \theta_{Q}^{*},X^{(V)})}{\partial\theta_{Q}^{*}} \\
%                &\approx  \epsilon^{T} (\frac{\partial \cL_{R}(\theta_{FP}, \theta_{Q}^{*},X^{(V)})}{\partial\theta_{Q}^{*}} - \frac{\partial \cL_{R}(\theta_{FP}, \theta_{Q},X^{(V)})}{\partial\theta_{Q}} + \\
%                & \quad \frac{\partial \cL_{R}(\theta_{FP}, \theta_{Q},X^{(V)})}{\partial\theta_{Q}} ) \\
%                &\approx  \epsilon^{T} (\frac{\partial \cL_{R}(\theta_{FP}, \theta_{Q}^{*},X^{(V)})}{\partial\theta_{Q}^{*}} - \frac{\partial \cL_{R}(\theta_{FP}, \theta_{Q},X^{(V)})}{\partial\theta_{Q}}) + \\
%                & \quad \epsilon^{T} \frac{\partial \cL_{R}(\theta_{FP}, \theta_{Q},X^{(V)})}{\partial\theta_{Q}}  \\
%          \end{aligned}
%    \end{equation}
\begin{equation}
\small
        \label{eq:estimate_sam}
          \begin{aligned}[b]
                & \qquad \cL_{TOTAL}(\theta_{Q},\theta_{FP},X^{(V)},X^{(T)}) \\
                &\approx  \epsilon^{T} \triangledown_{\theta_Q^{*}} \cL_{R}(X^{(V)}) -\alpha\triangledown_{\theta_Q} \cL_{R}(X^{(T)})^{T} \triangledown_{\theta_Q} \cL_{R}(X^{(V)}),  
          \end{aligned}
    \end{equation}
where the second term $-\alpha\triangledown_{\theta_Q} \cL_{R}(X^{(T)})^{T} \triangledown_{\theta_Q} \cL_{R}(X^{(V)})$ is the matching of gradients of the reconstruction loss $\cL_R$ w.r.t. the model $\theta_Q$ when evaluating the synthetic training set $X^{(T)}$ and the real validation set $X^{(V)}$, respectively. 
%between the reconstruction losses for the training set $X^{(T)}$ and the validation set $X^{(V)}$. 
Replace $\epsilon$ with its closed-form solution in Eq.~(\ref{eq:eps_definition}), the \textit{first term} of $\cL_{TOTAL}(.)$ in~(\ref{eq:estimate_sam}) can be represented as:
\begin{equation}
 \small
        \label{eq:estimate_sam_step_2}
          \begin{aligned}[b]
                &  \rho \left(\frac{\triangledown_{\theta_Q^{*}} \cL_{R}(X^{(V)})}{\norm{\triangledown_{\theta_{Q}^{*}}\cL_{R}(X^{(V)})}}\right)^T  \triangledown_{\theta_Q^{*}} \cL_{R}(X^{(V)}) \\
                &=  \rho\norm{\triangledown_{\theta_{Q}^{*}}\cL_{R}(X^{(V)})} \\
                &=  \rho \norm{\triangledown_{\theta_Q^{*}} \cL_{R}(X^{(V)}) - \triangledown_{\theta_Q} \cL_{R}(X^{(V)}) \hspace{0.5em}+ \quad \triangledown_{\theta_Q} \cL_{R}(X^{(V)})}\\
                &\approx  \rho ||\explain{\triangledown_{\theta_Q^{*}} \cL_{R}(X^{(V)}) - \triangledown_{\theta_Q} \cL_{R}(X^{(V)})}{First-order Taylor} +  \triangledown_{\theta_Q} \cL_{R}(X^{(V)}) ||  \\
                &\approx \qquad \qquad  \rho \norm{\cH^{(\theta_{Q})} \delta_{\theta_{Q}}  \qquad \quad+  \triangledown_{\theta_Q} \cL_{R}(X^{(V)})},
          \end{aligned}
    \end{equation}
    %\end{multline}
\normalsize
%Using the approximation in Eq. \ref{eq:eps_approximate}, we can see that the second term  Eq. \ref{eq:estimate_sam} is independent w.r.t the training set $X^{(T)}$ that we want to generate. Therefore, we can discard this term and  optimize the SAM loss by minimizing only the first term:
%\begin{equation}
%        \label{eq:approximate_L_SAM_1}
%          \begin{aligned}[b]
%                & \qquad \arg \min_{X^{(T)}} \cL_{SAM}( \theta_{Q}^{*},\theta_{FP},X^{(V)}) \\
%                &\approx  \arg \min_{X^{(T)}} \rho \frac{\nabla_{\theta_{Q}}\cL_{R}(\theta_{Q},\theta_{FP},X^{(V)})}{||\nabla_{\theta_{Q}}\cL_{R}(\theta_{Q},\theta_{FP},X^{(V)}||} \frac{\partial^{2} \cL_{R}( \theta_{Q},\theta_{FP},X^{(V)})}{\partial^{2} \theta_{Q}} \\
%                & \quad (-\alpha \cJ(\theta_{Q},X^{(T)})) \\
%                &\approx \arg \min_{X^{(T)}} -|\cJ(\theta_{Q},X^{(V)})| \cH(\theta_{Q},X^{(V)})  \cJ(\theta_{Q},X^{(T)}) \\
%          \end{aligned}
%    \end{equation}
where $\cH^{(\theta_Q)}$ denotes the Hessian matrix over the model weights using the validation set $X^{(V)}$. The final row in~(\ref{eq:estimate_sam_step_2}) is the result of using the first-order Taylor expansion for $\triangledown_{\theta_Q^{*}} \cL_{R}(X^{(V)})$ around $\theta_Q$. From Eq. (\ref{eq:estimate_sam_step_2}), minimizing the first term of $\cL_{TOTAL}(.)$ in~(\ref{eq:estimate_sam}) is equivalent to minimizing the magnitude of $\cH^{(\theta_{Q})} \delta_{\theta_{Q}} +  \triangledown_{\theta_Q} \cL_{R}(X^{(V)})$. It is worth noting that $\triangledown_{\theta_Q} \cL_{R}(X^{(V)})$ and $\cH^{(\theta_{Q})}$ are independent from the calibration set $X^{(T)}$ that we want to generate.  %and optimize. %Suppose that the magnitude of  $\cH^{(\theta_{Q})} \delta_{\theta_{Q}}$ is fixed,  
The necessity condition for Eq. (\ref{eq:estimate_sam_step_2}) to be minimized is when $\cH^{(\theta_{Q})} \delta_{\theta_{Q}}$ and $\triangledown_{\theta_Q} \cL_{R}(X^{(V)})$ {have opposite directions}, % we ignore the magnitudes, does this raise concerns?
which is equivalent to optimize:
\begin{equation}
\small
        \label{eq:approximate_L_SAM_2}
          \begin{aligned}[b]
                & \quad \arg \min_{X^{(T)}} cos(\cH^{(\theta_{Q})} \delta_{\theta_{Q}}  , \triangledown_{\theta_Q} \cL_{R}(X^{(V)})) \\
                & \equiv \arg \max_{X^{(T)}}  cos(\cH^{(\theta_{Q})} \triangledown_{\theta_Q} \cL_{R}(X^{(T)}), \triangledown_{\theta_Q} \cL_{R}(X^{(V)})). 
        \end{aligned}
    \end{equation}
Estimating $\cH^{(\theta_{Q})}$ and $\triangledown_{\theta_Q}\cL_{R}(X^{(T)})$ for the whole model can be very expensive, therefore, we turn to utilize the Hessian matrix $\cH^{(\theta_{Q})}_{l}$ and the Jacobian vector $\triangledown_{\theta_Q}\cL_{R}(X^{(T)})_{l}$ of one layer $l$ instead. {As many quantization methods~\cite{LQ_Nets, Dong2019HAWQHA, AdaDFQ, Li2021BRECQPT, Genie} usually  keep the bit-width of the first convolutional layer and the last fully-connected layer higher than other layers in low-bit width setting~ (e.g., in 2/2 setting, they are still kept at 8 bits while other layers are quantized to 2 bits), %optimization based on the gradient of these two layers are more likely to make the most impact to the model's performance.
optimization based on the gradient of these two layers is likely to have more impact to the model's performance.} %the most sensitive layers of the quantized model are usually the first convolutional layer and the last fully-connected layer~\cite{XYZ}}, % we need to add refs here
Therefore, we choose to use the fully-connected layer for our estimation, as it usually has a far higher number of parameters (more influence) than the first convolutional layer and its  Jacobian matrix is easy to compute. For instance, for ResNet-18 architecture, the first convolutional layer has  $64\times7\times7\times3=9,408$ parameters, while the last fully connected layer has $512\times1000=512,000$ parameters. %Moreover, despite having higher number or parameters, the last layer of the network  is just a linear layer, its Jacobian matrix is less expensive and easier to compute as we show below.
    
    Let $\cA^{(T)}=\mathtt{f}(\theta_{Q},X^{(T)},L-1)$ and $\cA^{(V)}=\mathtt{f}(\theta_{Q},X^{(V)},L-1)$ denote respectively input for the fully-connected layer of the training and validation sets with size $N\times d$ where $N$ is the number of samples, and $\cW_{(L)}$ be the weight matrix of the fully-connected layer with size $d\times C$ where $C$ is the number of classes. We have:
    \begin{equation}
     \small
        \label{eq:Hessian}
          \begin{aligned}[b]
                \tf(\theta_{Q},X^{(T)},L) = \cA^{(T)}\cW_{(L)}   \\
                \tf(\theta_{Q},X^{(V)},L) = \cA^{(V)}\cW_{(L)}.
          \end{aligned}
    \end{equation}
    From Eq. (\ref{eq:Hessian}), the Jacobian matrix w.r.t the fully connected layer can be easily estimated as:
    \begin{equation}
    \small
        \label{eq:grad_fc}
          \begin{aligned}[b]
            % \triangledown_{\theta_Q}\cL_{R}(X^{(T)}) \approx \triangledown_{\theta_Q}\cL_{R}(X^{(T)})_{(L)} 
            %     =flatten\left({\cA^{(T)}}^T g^{(T)}\right)  \\
         \triangledown_{\theta_Q}\cL_{R}(X^{(T)})_{(L)} 
                =flatten\left({\cA^{(T)}}^T g^{(T)}\right),
          \end{aligned}
    \end{equation}
    where $g^{(T)}_{i} = \tf(\theta_{Q},\bx_i,L) - \tf(\theta_{FP},\bx_i,L)$. % is the gradient of the output from the last layer for the $i^{th}$ sample in the training set.
    As the fully-connected layer is just a linear layer, the Hessian matrix $\cH^{(\theta_{Q})}_{(L)}$ of this layer has the size $dC \times dC$ and can be represented be:
    \begin{equation}
     \small
        \label{eq:Hessian_0}
          \begin{aligned}[b]
                      \begin{pmatrix}
                        {\cA^{(V)}}^T \cA^{(V)}   & 0 & \dots & 0 \\
                        0 & {\cA^{(V)}}^T \cA^{(V)}   & \dots & 0 \\
                        \vdots & \vdots & \ddots & \vdots \\
                        0 & 0 & \dots & {\cA^{(V)}}^T \cA^{(V)}
                      \end{pmatrix}
          \end{aligned}
    \end{equation}
    We can see that the Hessian matrix in (\ref{eq:Hessian_0}) has a positive diagonal, its diagonal elements are periodically overlapped and most of its off-diagonal elements are 0. Note that this matrix is independent of the training set $X^{(T)}$ that we want to generate. Similar to AdaRound ~\cite{Nagel2020UpOD}, we assume that $\cH^{(\theta_{Q})}_{(L)}$ is a diagonal matrix with the same main diagonal value:
    \begin{equation}
        \label{eq:Hessian_diagonal}
          \begin{aligned}[b]
               \cH^{(\theta_{Q})}_{(L)} = cI,                  
          \end{aligned}
    \end{equation}
    where $c$ is some constant, and $I$ denotes the identity matrix. Then  (\ref{eq:approximate_L_SAM_2}) is equivalent to:
    \begin{equation}
    \label{eq:approximate_L_SAM_4}
          \begin{aligned}[b]
                 \arg \max_{X^{(T)}} cos(\triangledown_{\theta_Q}\cL_{R}(X^{(T)}),  \triangledown_{\theta_Q}\cL_{R}(X^{(V)})). 
          \end{aligned}
    \end{equation}    
{As (\ref{eq:approximate_L_SAM_4}) and the second term of $\cL_{TOTAL}(.)$ in Eq.~(\ref{eq:estimate_sam}) correlate, optimizing~(\ref{eq:approximate_L_SAM_4}) will also {partly} optimize the second term  in Eq.~(\ref{eq:estimate_sam}). %, therefore we can summarize the optimization of Eq.~(\ref{eq:estimate_sam}) in Theorem (\ref{theorem:gm}). 

%\begin{equation}
%        \label{eq:gradient_matching_connect}
%          \begin{aligned}[b]
%          \arg \max_{\{x^{(T)}_i\}_{i=1}^T} \triangledown_{\theta_Q}\cL_{R}(X^{(T)})^T\triangledown_{\theta_Q}\cL_{R}(X^{(V)})
%          \end{aligned}
%    \end{equation}

 %

%  \begin{theorem}
%  \label{theorem:gm}
% Under the assumptions that the Hessian matrix of the last layer of deep neuron network is a diagonal with  matrix with, the minimization of total of the sharpness loss $\cL_{SAM}(.)$ and reconstruction loss $\cL_{R}(.)$  can be achieved by maximizing the gradient matching between the calibrated set and the validation set:
%   \begin{equation}
%          \label{eq:gradient_matching_connect}
%            \begin{aligned}[b]
%            \arg \max_{\{x^{(T)}_i\}_{i=1}^T} cos(\triangledown_{\theta_Q}\cL_{R}(X^{(T)})^T\triangledown_{\theta_Q}\cL_{R}(X^{(V)}))
%            \end{aligned}
%      \end{equation}
% \end{theorem}

\subsection{Gradient matching without a  validation set}

\begin{algorithm}[t]
%\small
\caption{SA zero-shot quantization.}
    \label{alg:full_alg_1}
    \begin{algorithmic}[1]
              
    \STATE \textbf{Train}{($\theta_{FP}$,$\cG$,$T$,$N_{w}$,$N_{g}$,$N_{q}$)}.
    
%\STATE{\(\theta_{Q}\): Quantization model weights.}
\STATE {\(\theta_{FP}\): The full-precision model.}\\
%\STATE {\(\bz\): The embedding set.}
\STATE {\(\cG\): The generator.}           \\ 
\STATE {\(T\): Number of images we want to generate.} \\
\STATE {\(N_{w}\): Number of warm-up iterations.} \\
\STATE {\(N_{g}\): Number of iterations for data optimization.}\\
\STATE {\(N_{q}\): Number of iterations for model quantization.}

\STATE {Initialize $\theta_{Q}$ from $\theta_{FP}$}.  
\STATE Initialize $\cG$ and $Z \sim  \mathcal{N}(\textbf{0},\textbf{I})$ %(i.e., embeddings -- inputs for the generator).
\STATE Warm up   $\theta_{Q}$, $\cG$, $z$ with BN loss (\ref{eq:bn_loss}) for $N_{w}$ iterations.

\FOR {$j = 1$ to $N_{g}$}
\FOR {$i = 1$ to $T$}
\STATE $\epsilon^{(\cN)}_{i} = \nu \frac{\triangledown_{\bz_i} \mathcal{D}(\triangledown_{\theta_Q}\cL_{R}(\cG(\bz_i)),\triangledown_{\theta_Q}\cL_{R}(\cG(\bz_i)))}{\norm{\triangledown_{\bz_i} \mathcal{D}(\triangledown_{\theta_Q}\cL_{R}(\cG(\bz_i)),\triangledown_{\theta_Q}\cL_{R}(\cG(\bz_i)))} } $ \\
\STATE $\bz_i = \bz_i - \gamma \partial\frac{\mathcal{L}_{FINAL} }{\bz_i}$ 
\STATE $\cG = \cG - \gamma \partial\frac{\mathcal{L}_{FINAL} }{\cG}$ 

\ENDFOR
\ENDFOR
\STATE Get the dataset $X^{(T)} \coloneqq \{\bx^{(T)}_i | \bx^{(T)}_i = \cG(\bz_i)\}$.
\FOR {$t = 1$ to $N_{q}$}
\STATE $\theta_{Q} = \theta_{Q} - \alpha_{\theta} \frac{\partial \mathcal{L}_{R}(\theta_{Q},\theta_{FP},X^{(T)})}{ \partial \theta_{Q}}$.
\ENDFOR           
\STATE \textbf{return}  $\theta_{Q}$.
\end{algorithmic} 
\end{algorithm}

From~(\ref{eq:approximate_L_SAM_4}), intuitively, we can see that the small set $X^{(T)}$  that we want to generate for calibration can decrease the SAM loss of the model over the validation set $X^{(V)}$ when the gradient of the model when evaluating on $X^{(T)}$  matches with that when evaluating on $X^{(V)}$. In summary, there is a strong correlation between the minimization of the SAM loss and the matching of the gradients of the reconstruction loss $\cL_{R}(.)$ w.r.t. $\theta_Q$ evaluated on the calibrated set and the validation set. Unfortunately, this objective is still hard to optimize because we do not have a large validation set $X^{(V)}$. Thus, we propose a novel approach to approximate (\ref{eq:approximate_L_SAM_4}).
Our idea is that if the training set is diverse enough,i.e., it spans over the data space, then for each sample $\bx^{(T)}_i$ that we want to generate, we just need to optimize its gradient to match with the gradients of samples in its neighborhood. In other word, we want to generate a training set $X^{(T)}$ such that samples of $X^{(T)}$ are diverse and each sample of $X^{(T)}$ maximizes the gradient matching with its neighbors.  
To this end, we introduce an approach similar to SAM. We want to minimize the difference between the gradients of each generated sample and a sample in its neighbor with the highest gradient dissimilarity. Specifically, given a generated training set $X^{(T)} \coloneqq \{\bx^{(T)}_i | \bx^{(T)}_i = \cG(\bz_i)\}$, with $\cG(.)$ is a generator and $Z\coloneqq \{ \bz_i\}_{i=1}^{|X^{(T)}|}$  is the set of embedding vectors to generate $X^{(T)}$, we want to estimate the perturbation $\epsilon^{(\cN)} = \{\epsilon^{(\cN)}_{i}\}_{i=1}^{|X^{(T)}|}$ such that ${\bx^*}^{(T)}_i = \cG(\bz_i +\epsilon^{(\cN)}_{i})$ is the sample with the least similarity in gradient within the neighborhood of  the original samples $\bx^{(T)}_i$: 
    \begin{equation}
        \small
        \label{eq:approximate_epsilon_sample}
          \begin{aligned}[b] & \epsilon^{(\cN)}_{i} =  \arg \max_{\epsilon^{(\cN)}_{i}} \mathcal{D}(\triangledown_{\theta_Q}\cL_{R}(\cG(\bz_i)),\triangledown_{\theta_Q}\cL_{R}(\cG(\bz_i+\epsilon^{(\cN)}_{i}) ))
          \\
           &\text{ s.t: } ||\epsilon^{(\cN)}_{i}|| \leq \nu,
          \end{aligned}
    \end{equation}
    where $\mathcal{D}(.)$ is a loss that measures the difference in gradients, assumed to be the cosine similarity distance. Following~\cite{FirstSam}, the closed-form solution for $\epsilon^{(\cN)}_{i}$ is:%, we use a similar closed-form solution from the traditional SAM~\cite{FirstSam} approach:
    \vspace{-0.2cm}
    \begin{equation}
     \small
\label{eq:approximate_epsilon_formula}
          \begin{aligned}[b] & \epsilon^{(\cN)}_{i} = \nu \frac{\triangledown_{\bz_i} \mathcal{D}(\triangledown_{\theta_Q}\cL_{R}(\cG(\bz_i)),\triangledown_{\theta_Q}\cL_{R}(\cG(\bz_i)))}{\norm{\triangledown_{\bz_i} \mathcal{D}(\triangledown_{\theta_Q}\cL_{R}(\cG(\bz_i)),\triangledown_{\theta_Q}\cL_{R}(\cG(\bz_i)))} }. 
          \end{aligned}
    \end{equation}
    Given $\epsilon^{(\cN)}$, we can estimate the sample that has its gradient matches with the gradients of its neighborhoods:
\begin{equation}
    \small
\label{eq:optimize_sample_grad}
\begin{aligned}[b] 
 \quad  \bz_i,\cG & = \arg \min_{\bz_i,\cG} \cL_{GRAD}(\theta_{Q},\theta_{FP},X^{(T)}) \\
 &= \arg \min_{\bz_i,\cG} \mathcal{D}(\triangledown_{\theta_Q}\cL_{R}(\cG(\bz_i)),\triangledown_{\theta_Q}\cL_{R}(\cG(\bz_i +\epsilon_i^{(\cN)}))) \\
 &\forall i=1,2,..,|X^{(T)}|, 
\end{aligned}  
\end{equation}
%The above gradient matching loss  only makes sense if the calibration set is diverse enough. 
The above gradient matching loss will encourage (\ref{eq:approximate_L_SAM_4}) if the calibration set $X^{(T)}$ is diverse enough. 
%, \blue{such that each sample in the validation set $X^{(V)}$ is within the neighborhood of at least one sample from the training set $X^{(T)}$}, 
Therefore, after normalizing $\triangledown_{\theta_Q}\cL_{R}(X^{(T)})$ to be unit vectors,  we add a term to encourage the diversity of the calibration set:
\begin{equation}       
 \small
\label{eq:optimize_diversity}
          \begin{aligned}[b] 
          & \quad \cL_{DIVERSE}(\theta_{Q},X^{(T)}) \\ 
          &=  \sum_{i, j, i\neq j} \max (0,abs({\triangledown_{\theta_Q} \cL_{R}^{(i)}}^T \triangledown_{\theta_Q} \cL_{R}^{(j)}) - \zeta),  \\
          \end{aligned} 
    \end{equation}
    \vspace{-0.1cm}
where $abs(.)$ denotes the absolute function;   $\triangledown_{\theta_Q}\cL_{R}^{(j)}=\triangledown_{\theta_Q}\cL_{R}(\bx_j)$ represents the gradient of the reconstruction loss w.r.t. $\theta_Q$ for sample $\bx_j$; $\zeta$ is a small positive threshold. %This diversity loss penalizes samples that have gradient similarity above a small positive threshold $\zeta$. 
This loss encourages gradients of samples $x_i$ and $x_j$ ($i\neq j$) to be orthogonal when $\zeta$ is close to 0. Consequently, it encourages the calibration set to be diverse.

\paragraph{Verification of the gradient matching on real data.}
To verify the effectiveness of the gradient matching for the quantization process, we conduct experiments using the gradient matching on real images. Given a set of real images, we compare the performance of the quantized model calibrated over a randomly extracted subset of data and a subset of data that minimizes the gradient matching loss in~(\ref{eq:approximate_L_SAM_4}). 
The results presented in Table~\ref{tab:performance_gain_random} show that the gradient matching loss   (\ref{eq:approximate_L_SAM_4}) consistently improves the performance of the quantized model. %, regardless of the calibration set size.
The improvements are more significant when the number of samples is small. 
\vspace{-0.5cm}
\begin{table}[h]
\caption{The comparative performance when quantizing the ResNet-18 model using the real data randomly selected and the real data selected by~(\ref{eq:approximate_L_SAM_4}).
%gain when the real data are used for the quantization with the ResNet-18 model using gradient matching loss (\ref{eq:approximate_L_SAM_4}).% compared to the random selection of data.
}
%\vspace{0.1 cm}
\label{tab:performance_gain_random}
  \centering
  \begin{tabular}{l|cccc}
    \hline
    Num. Images & {32} & {64} & {128} &    {256} \\
    \hline
    Random  & {32.67} & {43.15} & {49.45} & {53.64}  \\
    SADAG (Ours) & {35.46} & {44.05} & {50.30} & {54.01}  \\
    \hline
  \end{tabular}
  \vspace{-4mm}
\end{table}

\subsection{Final algorithm}

Besides the sharpness of the model, we also want the generated samples to follow the distribution of original data. Particularly, we encourage $X^{(T)}$ %to contain knowledge of the real data by encouraging them 
to have similar BN statistics stored in the BN layers of the full-precision model $\theta_{FP}$, by introducing the BN loss $L_{BN}$: 
\begin{equation}
\small
    \label{eq:bn_loss}
          \begin{aligned}[b]
          \cL_{BN}(\theta_{FP},X^{(T)}) &= \sum_{j=1}^{L} (||\mu_{j}^{(s)} - \mu_{j}||^{2}+ ||\sigma_{j}^{(s)} - \sigma_{j}||^{2}),
          \end{aligned}
    \end{equation}
where $\mu_{j}^{(s)}$ and $\mu_{j}$ are respectively the mean output values of the synthetic dataset $X^{(T)}$ from the full-precision model at the $j^{th}$ layer and the BatchNorm statistic of the full-precision model from the same layer, while $\sigma_{j}^{(s)}$ and $\sigma_{j}$ are the corresponding standard deviations. % at intermediate layer $j^{th}$ of the synthetic dataset $X^{(T)}$ and those from the pretrained model.
%
%While using Eq. (\ref{eq:gradient_matching_connect}) for optimization is good enough, we can also exactly optimize the calibrated dataset with Eq. (\ref{eq:approximate_L_SAM_3}) without the need for any relaxation. Using the definition of Jacobian vector $\triangledown_{\theta_Q}\cL_{R}(\bx^{(T)}_i) = flatten\left(g_{T}\cX_{(T)}^T\right)$ in Eq. (\ref{eq:grad_fc}), and combine this with Eq. (\ref{eq:Hessian_0}) and (\ref{eq:approximate_L_SAM_3}), we can transform (\ref{eq:approximate_L_SAM_3}) as:
%    \begin{equation}
%        \label{eq:1_layer_approximate_3}
%        \begin{aligned}[b]
%            & \quad \arg \max_{X^{(T)}} \sum_{i} g_{T}^{(i)}\cX_{(T)}^T \left( \cX_{(V)}  \cX_{(V)}^T\right) \cX_{(V)} g_{V}^{(i)} \\
%            & = \arg \max_{X^{(T)}} \sum_{i} g_{T}^{(i)}g_{V}^{(i)T} \odot \cX_{(T)}^T \left(\cX_{(V)}  \cX_{(V)}^T\right) \cX_{(V)} 
%        \end{aligned}
%    \end{equation}
%In fact, Eq. (\ref{eq:1_layer_approximate_3}) is the same as $\mathcal{D}(\triangledown_{\theta_Q}\cL_{R}(\bx^{(T)}_i),\triangledown_{\theta_Q}\cL_{R}(\bx^{(T)}_i + \epsilon^{(\cN)}_{i}))$, but we have an extra term $\left( \cX_{(V)}  \cX_{(V)}^T\right)$ in the middle. Need to note that there is difference between the $\cX_{(V)}$ in the middle term $\cH=\cX_{(V)}  \cX_{(V)}^T$ and the final $\cX_{(V)}$, as the first one represent the Hessian matrix of the last layer, while the second one are neighbor samples of current synthetic calibrated set. 
Initially, we need to warm up the calibrated set $X^{(T)}$ using a  data generation method. % to make sure that they represent distribution of original data. 
%We then use these warmup data's feature from the full-precision model $\cX_{(T)}$ to estimate the middle term $\cH=\cX_{(T)}  \cX_{(T)}^T$ and keep it fixed for the rest of the training process. 
After the warm-up stage, we acquire the final calibration set $X^{(T)}$ by minimizing the loss in Eq.~(\ref{eq:final_loss}) over the warm-up data. Finally, this newly generated training set will be used to calibrate the current quantized model.
\begin{equation}
\small
    \label{eq:final_loss}
          \begin{aligned}[b]
          & \cL_{FINAL}(\theta_{FP},X^{(T)}) = \cL_{BN}(\theta_{FP},X^{(T)}) + \\ &\qquad \lambda_1\cL_{DIVERSE}(\theta_{Q},X^{(T)}) + \lambda_2 \cL_{GRAD}(\theta_{Q},\theta_{FP},X^{(T)}).
          \end{aligned}
    \end{equation}
    \normalsize
    
    Algorithm~\ref{alg:full_alg_1} gives the overall algorithm of our proposed method.
    \vspace{-0.3cm}

\section{Experiments}
\label{sec:experiment}
\subsection{Experimental setup}
\paragraph{Dataset and network architecture.} 
We evaluate our approach on %CIFAR-10~\cite{CIFAR}, 
CIFAR-100~\cite{CIFAR} and ImageNet~\cite{imagenet} datasets, which are commonly utilized for zero-shot quantization.  Following the settings in~\cite{AdaDFQ, HAST, qimera}, we evaluate our proposed method using ResNet-20~\cite{resnet} model for CIFAR-100 dataset. For the ImageNet dataset, we validate our proposed approach using ResNet-18~\cite{resnet}, ResNet-50 ~\cite{resnet}, and MobileNetV2~\cite{Mobilenetv2} architectures.
\paragraph{Quantization setting.}
Following Genie~\cite{Genie}, we adopt the uniform quantization scheme with adaptive rounding approach~\cite{Nagel2020UpOD}, as elaborated in \cref{subsec:quantization_scheme}. %Throughout the experimentation process, 
In our experiments, %we maintain a consistent configuration whereby 
the bit-widths of the first layer and the last layer are fixed at 8 bits, which is similar to recent SOTA methods for PTQ~\cite{Genie,Li2021BRECQPT}. Following BRECQ ~\cite{Li2021BRECQPT} and Genie~\cite{Genie}, we also set the activation bit-widths of the second layer and the last layer to 8 bits. The remaining weight and activation bit-widths of other layers follow the specifications in the corresponding experimental setups. %The remaining weight and activation bit-widths adhere to the specifications outlined in the respective experimental setups.

\paragraph{Implementation details.}
For the initial warm up of the synthetic dataset, we adopt a generator and a set of embedding vectors with 256 dimensions for each mini-batch of images that we generate, similar to the Genie model~\cite{Genie}. We use the same quantization setting as BRECQ~\cite{Li2021BRECQPT}. The learning rates of the generator and embedding are initially set at  0.1 and 0.01, respectively. 
We adopt the Adam optimizer~\cite{Kingma2014AdamAM} for both generator and data embedding, but utilize different schedulers for them, i.e., the ExponentialLR scheduler and ReduceLRonPlateau scheduler are used for scheduling the learning rates of the generator and the embeddings, respectively. %For the generator, we schedule its learning rate with ExponentialLR scheduler, while ReduceLRonPlateau scheduler is used for the embedding.  
Across all experiments, the batch size for the data generation process is set to 128, while in the quantization step, we keep the batch size at 32. {The threshold $\zeta$ in Eq.~(\ref{eq:optimize_diversity}) is set to 0 or 0.1}. %while the magnitude of upper bound for embedding perturbation $\nu$ is set to 2.
{The radius $\nu$ in Eq.~(\ref{eq:approximate_epsilon_sample}) for the embedding perturbation is set to 2.} To demonstrate the effectiveness of our method, we compare our model's performance with the recent state-of-the-art models on different low-bit ZSQ settings. {Following previous works~\cite{Genie,AdaDFQ}} and combined with an additional extreme low-bit setting (2/2),  we use a total of 4 different quantization settings, including 2/2, 3/3, 2/4, and 4/4 bit-width for ImageNet experiments. On the other hand, for CIFAR-100, we report the results of 3/3 and 4/4 bit-width settings. 
Our model is then compared with recent prominent zero-shot quantization models, i.e., Qimera~\cite{qimera}, AdaSG~\cite{AdaSG}, IntraQ~\cite{IntraQ}, AdaDFQ~\cite{AdaDFQ} and Genie~\cite{Genie}. Regarding the number of generated images, we generate a total of 1,024 images for a fair comparison with Genie~\cite{Genie}. %, \blue{while other methods either generate a minibatch per iteration~\cite{AdaDFQ} or use 5,120 samples~\cite{IntraQ}.}
% The implementation details and the values of the hyperparameters are provided in the supplementary material.
\subsection{Comparison with the state of art}
Table~\ref{tab:CIFAR100} represents the comparative results of our method SADAG and other state-of-the-art (SOTA) methods when evaluated on the CIFAR-100 dataset. The results of~\cite{qimera,AdaSG,IntraQ,AdaDFQ} are cited from~\cite{AdaDFQ}, while for Genie~\cite{Genie}, we use their official released code and adapt it for the ResNet-20 model. 
%The results of our method and other SOTA models for CIFAR-100 dataset are presented in Table \ref{tab:CIFAR100}.
The results confirm the superior performance of the proposed SADAG over the state of the art. 
%Overall, our method SADAG outperforms other methods by a noticeable margin on both bit-width settings. 
Comparing to Genie~\cite{Genie}, the current SOTA model in ZSQ, our method improves over Genie 0.69\% and 0.76\% for the 3/3 and 4/4 settings, respectively. %Moreover, our method seem to work better in extremely low-bit setting (2/2). 

\begin{table}[t]
  \caption{Comparisons of Top-1 classification accuracy (\%) with the state of the art on CIFAR-100 dataset.}
  \vspace{0.1 cm}
  \label{tab:CIFAR100}
\begin{tabular}{c|c|c}
\hline
\multirow{2}{4em}{Method} & {Bit-width} & {ResNet-20} \\ 
&  (W/A) & (FP: 70.33) \\
\hline
 Qimera~\cite{qimera} & \multirow{6}{*}{3/3} & 46.13 \\
%AdaQuant~\cite{AdaQuant} & 2/2 & - & - & - \\
 AdaSG ~\cite{AdaSG} &  & 52.76   \\
 IntraQ~\cite{IntraQ} &  &  - \\

  AdaDFQ~\cite{AdaDFQ} &  &  52.74 \\
 Genie~\cite{Genie} &  &  65.25  \\
 SADAG (Ours) &  & \textbf{65.94}  \\ 
\hline
 Qimera~\cite{qimera} & \multirow{6}{*}{4/4} & 65.10 \\
%AdaQuant~\cite{AdaQuant} & 2/2 & - & - & - \\
 AdaSG ~\cite{AdaSG} &  &  66.42  \\
 IntraQ~\cite{IntraQ} &  & 64.98 \\

 AdaDFQ~\cite{AdaDFQ} &  & 66.81 \\
 Genie~\cite{Genie} &  &  68.35  \\
 SADAG (Ours) &  &  \textbf{69.11} \\
\hline
\end{tabular}
\vspace{-0.5cm}
\end{table}
%On the other hand, Table \ref{tab:imagenet} shows the results of our model compared with the current SOTA models in zero-shot quantization for ImageNet benchmark. 
Table~\ref{tab:imagenet} presents the comparative results of our method SADAG and other state-of-the art methods when evaluated on the ImageNet dataset. %It is worth noting that for the very low-bit setting 2/2, none of the compared methods report results for that setting. 
It is worth noting that Genie~\cite{Genie} -- the SOTA method for ZSQ does not report results for the 2/2 and 3/3 settings. We use their official released code to produce Genie's results for those settings. 
Regarding other results of the competitors, except for the 2/4 setting of Genie with the MobileNetV2 architecture which we produce it result by using the Genie's released code\footnote{While we are able to reproduce other results of Genie, we are unable to reproduce its result for the 2/4 setting with MobileNetV2. There is a large gap between the reported number in their paper and the reproduced result, i.e., 53.38 vs. 51.47.}, they are cited from~\cite{AdaDFQ,Genie}.

%All baseline results are taken from the original papers, with the exception of the 2/2 bit-width setting where we have to run their provided code repositories to get the results, and the special case for MobileNetV2 experiment of Genie~\cite{Genie} where we try to replicate their result with their code. 
The results in Table~\ref{tab:imagenet} show that our method SADAG consistently outperforms previous approaches, including the current best method Genie~\cite{Genie} on all bit-width settings and all considered model architectures, which confirms the effectiveness of the proposed sharpness-aware data generation approach. The improvements of our method over Genie are more clear in the 2/2 setting, i.e., the improvements are 0.77\%, 0.74\%, and 1.08\% for the ResNet-18, ResNet-50, and MobileNetV2, respectively. 

%, which confirms the effectiveness of the proposed sharpness-aware data generation approach. Compared to the current best model (Genie~\cite{Genie}), our method outperforms 0.2\%-0.8\% across most benchmarks and show competitive performance for the 4/4 setting. Similar to CIFAR-100 experiments, there is also bigger margin improvement in low-bit experiments.

\begin{table*}[t]
\centering
\caption{Comparisons of Top-1 classification accuracy (\%) with the state of the art on ImageNet dataset. The result denoted with (*) is reproduced using the official released code of the corresponding paper.}
 %\vspace{0.1cm}
  \label{tab:imagenet}
\begin{tabular}{l|c|cccc}
\hline
{Method} & {Bit-width (W/A)} & {ResNet-18} & {ResNet-50} & {MobileNetV2} \\ \hline \hline
 & Full precision & {71.01} & {76.63} & {72.20} \\
\hline
%Qimera~\cite{qimera} & \multirow{6}{*}{2/2} & - & - & - \\
%AdaSG ~\cite{AdaSG} &  & 0.124 &-  &-  \\
% IntraQ~\cite{IntraQ} &  & 0.161 &-  & - \\

%  AdaDFQ~\cite{AdaDFQ} &  & 0.170 & -  &-  \\
 Genie~\cite{Genie} & \multirow{2}{*}{2/2} & 53.74 & 56.81 & 11.93 \\
  SADAG (Ours) &  & \textbf{54.51} & \textbf{57.55} & \textbf{13.01} \\
\hline

%Qimera~\cite{qimera} & \multirow{6}{*}{2/4} & - & - & - \\
%AdaQuant~\cite{AdaQuant} & 2/2 & - & - & - \\
%AdaSG ~\cite{AdaSG} &  &-  &-  & - \\
% IntraQ~\cite{IntraQ} &  &-  &-  & - \\
 % AdaDFQ~\cite{AdaDFQ} &  &-  & - & - \\
 Genie~\cite{Genie} &\multirow{2}{*}{2/4}   & 65.10 & 69.99 & $51.47^{*}$  \\
  SADAG (Ours) &  & \textbf{65.25} & \textbf{70.52} & \textbf{51.89}\\
\hline

Qimera~\cite{qimera} & \multirow{6}{*}{3/3} & 1.17 & - & - \\
%AdaQuant~\cite{AdaQuant} & 2/2 & - & - & - \\
AdaSG ~\cite{AdaSG} &  & 37.04 & 16.98 & 26.90 \\
 IntraQ~\cite{IntraQ} &  & - & - & - \\

  AdaDFQ~\cite{AdaDFQ} &  & 38.10 & 17.63 & 28.99 \\
 Genie~\cite{Genie} &  & 66.89 & 72.54 & 55.31  \\
  SADAG (Ours) &  & \textbf{67.10} & \textbf{72.62}& \textbf{56.02} \\
\hline

Qimera~\cite{qimera} & \multirow{6}{*}{4/4} & 63.84 & 66.25 & 61.62 \\
%AdaQuant~\cite{AdaQuant} & 2/2 & - & - & - \\
AdaSG ~\cite{AdaSG} &  & 66.50 & 68.58 &  65.15 \\
 IntraQ~\cite{IntraQ} &  & 66.47 & - & 65.10 \\

  AdaDFQ~\cite{AdaDFQ} &  & 66.53 &  68.38 & 65.41 \\
 Genie~\cite{Genie} &  & 69.66 & 75.59 & 68.38 \\
  SADAG (Ours) &  & \textbf{69.72} & \textbf{75.7} &  \textbf{68.54} \\
\hline

\end{tabular}
\end{table*}

\subsection{Visualization and ablation studies}
%\subsubsection{Comparison with gradient matching using all model's weight}
\begin{figure*}[t]
        % \vspace{-0.5em}
        \centering
         \includegraphics[width=1\textwidth]
         {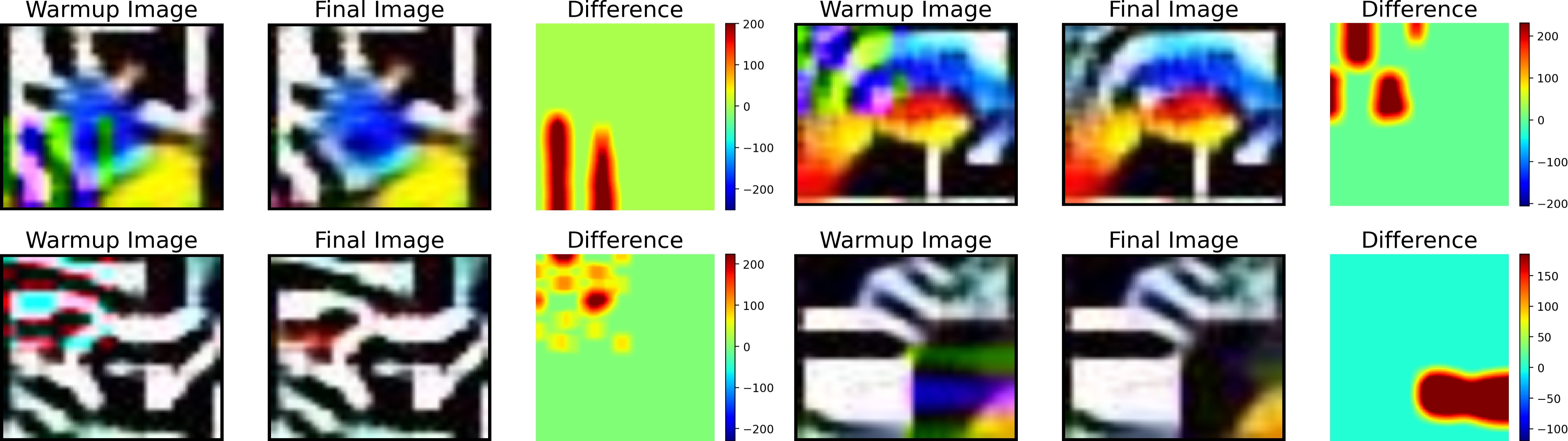} 
        \vspace{-0.5cm}
        \caption{The warm-up images and the corresponding images generated by our proposed method SADAG, and the corresponding heat maps of their differences. 
        %The color bars on the right-hand side denote the differences between pixel values of the warm-up and generated images.
        }
    \label{fig:generated_images}
    \vspace{-5mm}
    \end{figure*}
\subsubsection{Visualization}
In Figure \ref{fig:generated_images}, we present several examples of the warm-up images and the corresponding synthetic images generated by our proposed method. %that our framework generate at each stage of the process. 
The images for visualization are taken from the setting 3/3 with CIFAR-100 dataset with ResNet-20 model. In each corner of the figure, we show three different images that respectively represent the images after the warm-up stage (i.e., warm-up images), the final synthetic images, and the heat map demonstrating the pixel value differences between them. Because the BatchNorm loss converges very fast, the semantics of the images do not change much after the warm-up stage. %However, the heat map show there are differences in the pixel values between 
{However, we can observe that our method appears to make the images smoother compared to those in the warm-up stage, i.e., there is less variability in color in the generated images compared to the warm-up images.}
%, as the algorithm encourages the images to align their semantics with samples in their neighborhood. As a result, the synthetic images will move toward a more confident and higher density area in the data space.}

\subsubsection{Impact of the number of generated images}
%To investigate the effectiveness of our method when the size of the calibration set varies, 
We conduct an ablation study about the model's performance gain with different numbers of generated images, ranging from 128 to 1024. We use ResNet-18 architecture and the ImageNet dataset for evaluation, with the 2/2 bit-width for this experiment. The results are presented in Table~\ref{tab:ablatant_num_sample}. 
The results show that increasing the number generated images improves the model's performance. However, the performance gain is smaller when the number of images increase, e.g., for the proposed method, the performance gains are 3.71\% and 1.94\% when increasing the number of images from 128 to 256, and from 256 to 512, respectively. In addition, the results show that the proposed method consistently outperforms over Genie~\cite{Genie}. 
\begin{table}[!t]
  \caption{Comparative results between our method and Genie~\cite{Genie} with different numbers of generated images.}
   \vspace{0.1cm}
  \label{tab:ablatant_num_sample}
  \centering
  \begin{tabular}{l|cccc}
    \hline
    Num. Images & {128} & {256} & {512} & {1024} \\
    \hline
    Genie & {47.17} & {50.46} & {52.79} &{53.74} \\

    SADAG (Ours) & {47.54} & {51.25} & {53.19} & {54.51}  \\
    \hline
  \end{tabular}
  \vspace{-5mm}
\end{table}
%
%Based on the result, we can see that just by adding sharpness to the synthetic image, our method improves the performance regardless of the size of the calibration set, at a consistent rate.
\subsubsection{Impact of the loss factors $\lambda_1$ and $\lambda_2$}
%Another major hyperparameter that determines our model's performance is the contribution  ratio of each loss component, as denoted in Eq. (\ref{eq:final_loss}) by $\lambda_1$ and $\lambda_2$. Therefore, 
We conduct an ablation study to assess the impact of the hyperparameters $\lambda_1$ and $\lambda_2$ in Eq.~(\ref{eq:final_loss}). 
Specifically, we respectively keep one of the two variables $\lambda_1$ and $\lambda_2$ fixed at 1, while varying the other with five different values $0, 0.5, 1, 2, 5$. The experiments are conducted with ResNet-18 with the 2/2 setting and the number of generated images is 1024 for each experiment. The results are presented in Tables \ref{tab:lambda1_param} and \ref{tab:lambda2_param}. 
 %For simplicity, we use the same ResNet-18 architecture on ImageNet dataset and generate 128 samples for each experiment. 
%\blue{The results show that the $\cL_{GRAD}(.)$ loss is more sensitive than the $\cL_{DIVERSE}(.)$ loss. In addition, the model's performance degrades  with larger or smaller $\lambda_1$ and $\lambda_2$. Therefore, we simply keep both of them at value 1.}
%{The results show that the $\cL_{DIVERSE}(.)$ loss is more sensitive than the $\cL_{GRAD}(.)$ loss when $\lambda_1$ and $\lambda_2$ are small.} 
As we can see, the model's performance degrades  with larger or smaller $\lambda_1$ and $\lambda_2$. Therefore, we simply keep both of them at value 1.
\begin{table}[t]
  \caption{Change in performance w.r.t. $\lambda_1$ in Eq.~(\ref{eq:final_loss}).}
  \vspace{0.1cm}
  \label{tab:lambda1_param}
  \centering
  \begin{tabular}{l|ccccc}
    \hline
    $\lambda_1$ & 0 & 0.5 & {1} &{2} &{5}   \\
    \hline

    SADAG & 54.14 & 54.15 & {54.51} & {53.99} & {53.85}   \\
    \hline
  \end{tabular}
  \vspace{-3mm}
\end{table}

\begin{table}[t]
  \caption{Change in performance w.r.t. $\lambda_2$ in Eq.~(\ref{eq:final_loss}).}
  \vspace{0.1cm}
  \label{tab:lambda2_param}
  \centering
  \begin{tabular}{l|ccccc}
    \hline
    $\lambda_2$ & 0&  0.5& {1}  &{2}  &{5}  \\
    \hline

    SADAG &53.74  & 54.22& {54.51} &54.03  & {53.77}   \\
    \hline
  \end{tabular}
  \vspace{-5mm}
\end{table}

\subsubsection{Computation expense}
%\begin{table}[h!]
%  \caption{Comparative of execution time between out method and previous approach.}
%  \label{tab:ablatant_time_complex}
%  \centering
%  \begin{tabular}{l|ccccc}
%    \hline
%    Model & Genie & IntraQ & Qimera & AdaDFQ & Our \\
%%    \hline
 %   Time(h) & {2.5} & {6} & {6} &{6.5} &{5} \\

 %   \hline
 % \end{tabular}
%\%end{table}
Although the second-order objective in Eq.~(\ref{eq:L_SAM_expected}) is computationally intensive, we have successfully reduced the computational expense by approximating it with another first-order optimization in Eq. (\ref{eq:approximate_L_SAM_4}). Our proposed method operates at a speed that is approximately 1.5 times slower than Genie, which solely utilizes BatchNorm loss for optimization and is currently one of the fastest zero-shot quantization methods.  %The execution time is displayed in detail in Table. (\ref{tab:ablatant_time_complex}).

\section{Conclusion}
\label{sec:conclusion}
In this paper, we propose SADAG, a novel %a simple yet powerful 
approach for data-free quantization that takes into consideration the sharpness of the model calibrated on the synthetic dataset. We have elucidated the relationship between gradient matching between the training and validation sets and its influence on the calibrated model's sharpness on the validation set. 
%We have represented the correlation between the matching of gradient between the training and validation set, and how that may impact the performance of the calibrated model over the validation set. 
%
Our findings illustrate that enhancing the state-of-the-art generative data-free quantization can be accomplished without significant additional computational overhead or the necessity for any original data; simply by attending to the gradient of the neighborhood of each generated sample, indirect sharpness optimization over the hidden validation set is feasible. Extensive experimentation on two benchmark datasets %across various frameworks and datasets 
underscores SADAG's achievement of state-of-the-art performance. However, there still exist several limitations in our approach. A noticeable weakness of the framework is that it requires the relaxation for the Hessian matrix. As future
work, we will tackle this problem by integrating the model with some techniques to approximate the Hessian matrix. Another limitation is that our current method only takes into account the gradient of the final fully-connected layer instead of the whole model. Although accurate gradient matching approximation for the whole network is very computationally expensive, as potential improvement in the future, we can try to extend the method to match the gradient of several blocks instead of a single layer. %across diverse network architectures and datasets.
%Our results has demonstrated that the current state-of-the-art generative data-free quantization can be improved without incurring much additional computational expense nor the need of additional data, just by paying attention to the gradient of its neighborhood is enough for indirect sharpness optimization over the hidden validation set. Extensive experimental results on various framework and dataset shows that SADAG achieves state-of-the-art performance for many different network architecture and datasets.
\section*{Acknowledgements}
%We gratefully acknowledge the insightful comments and feedback from the reviewers, which greatly improved the quality of this paper. We are also thankful to Monash University for providing resources and facilities essential for the completion of this work.
Trung Le was partly supported by ARC DP23 grant DP230101176 and by the Air Force Office of Scientific Research under award number FA2386-23-1-4044.

\section*{Impact Statement}
This paper presents work whose goal is to advance the field of Machine Learning. There are many potential societal consequences of our work, none which we feel must be specifically highlighted here.

\bibliography{main}
\bibliographystyle{icml2024}

%%%%%%%%%%%%%%%%%%%%%%%%%%%%%%%%%%%%%%%%%%%%%%%%%%%%%%%%%%%%%%%%%%%%%%%%%%%%%%%
%%%%%%%%%%%%%%%%%%%%%%%%%%%%%%%%%%%%%%%%%%%%%%%%%%%%%%%%%%%%%%%%%%%%%%%%%%%%%%%
% APPENDIX
%%%%%%%%%%%%%%%%%%%%%%%%%%%%%%%%%%%%%%%%%%%%%%%%%%%%%%%%%%%%%%%%%%%%%%%%%%%%%%%
%%%%%%%%%%%%%%%%%%%%%%%%%%%%%%%%%%%%%%%%%%%%%%%%%%%%%%%%%%%%%%%%%%%%%%%%%%%%%%%
\newpage
\appendix
\onecolumn
\section{Appendix}
%%%%%%%%%%%%%%%%%%%%%%%%%%%%%%%%%%%%%%%%%%%%%%%%%%%%%%%%%%%%%%%%%%%%%%%%%%%%%%%
%%%%%%%%%%%%%%%%%%%%%%%%%%%%%%%%%%%%%%%%%%%%%%%%%%%%%%%%%%%%%%%%%%%%%%%%%%%%%%%

\end{document}